\pdfoutput=1

\documentclass[11pt]{article}

\usepackage[]{EMNLP2022}

\usepackage{times}
\usepackage{latexsym}

\usepackage[T1]{fontenc}

\usepackage[utf8]{inputenc}

\usepackage{microtype}

\usepackage{inconsolata}

\usepackage{amsmath}

\usepackage{microtype}
\usepackage{graphicx}
\usepackage{subfigure}
\usepackage{booktabs} 
\usepackage{latexsym}
\usepackage{threeparttable}
\usepackage{multirow}
\usepackage{amsfonts}
\usepackage{bbm}
\usepackage{bm}
\usepackage{bbold}
\usepackage{color}
\usepackage{enumitem}
\usepackage{geometry}

\title{Non-Parametric Domain Adaptation for End-to-End Speech Translation}

\author{
  Yichao Du$^{\ddag\sharp}$\thanks{\ \ Equal Contribution}, Weizhi Wang$^{\S *}$, Zhirui Zhang$^\natural$\thanks{\ \ Corresponding author}, Boxing Chen$^{\flat}$, Tong Xu$^{\ddag\sharp}$ \\ \textbf{Jun Xie$^{\flat}$ and Enhong Chen$^{\ddag\sharp\dag}$} \\
  $^{\ddag}$University of Science and Technology of China $^{\sharp}$State Key Laboratory of Cognitive Intelligence\\
  $^\S$University of California, Santa Barbara \ \ $^{\natural}$Tencent AI Lab \\
  $^{\flat}$Machine Intelligence Technology Lab, Alibaba DAMO Academy \\
  $^{\ddag\sharp}${duyichao@mail.ustc.edu.cn} \ $^{\ddag\sharp}${\{tongxu, cheneh\}@ustc.edu.cn}  $^{\S}${weizhiwang@ucsb.edu} \\ $^{\natural}${zrustc11@gmail.com} \  $^{\flat}${\{boxing.cbx, qingjing.xj\}@alibaba-inc.com}
}

\begin{document}
\maketitle
\begin{abstract}
The end-to-end speech translation~(E2E-ST) has received increasing attention due to the potential of its less error propagation, lower latency and fewer parameters.
However, the effectiveness of neural-based approaches to this task is severely limited by the available training corpus, especially for domain adaptation where in-domain triplet data is scarce or nonexistent. 
In this paper, we propose a novel non-parametric method that leverages in-domain text translation corpus to achieve domain adaptation for E2E-ST systems. 
To this end, we first incorporate an additional encoder into the pre-trained E2E-ST model to realize text translation modeling, based on which the decoder's output representations for text and speech translation tasks are unified by reducing the correspondent representation mismatch in available triplet training data.
During domain adaptation, a $k$-nearest-neighbor~($k$NN) classifier is introduced to produce the final translation distribution using the external datastore built by the domain-specific text translation corpus, while the universal output representation is adopted to perform a similarity search.
Experiments on the Europarl-ST benchmark demonstrate that when in-domain text translation data is involved only, our proposed approach significantly improves baseline by 12.82 BLEU on average in all translation directions, even outperforming the strong in-domain fine-tuning strategy.
\end{abstract}

\section{Introduction}
Speech translation~(ST), the task of automatically translating speech signals in a given language into text in another language, becomes a widely studied topic with the increasing demand for international communications.
Traditional ST systems cascade automatic speech recognition (ASR) and machine translation~(MT)~\cite{Ney1999SpeechTC, Sperber2017NeuralLM, Zhang2019LatticeTF, IranzoSnchez2020DirectSM,Machcek2021LostII}.
So far, various large-scale speech-translation datasets have been proposed, e.g., Libri-Trans~\cite{Kocabiyikoglu2018AugmentingLW}, MuST-C~\cite{Gangi2019MuSTCAM} and
CoVoST~\cite{Wang2020CoVoSTAD}.
With these large-scale annotations, building an end-to-end speech translation~(E2E-ST) system~\cite{Vila2018EndtoEndST,Liu2019EndtoEndST,Li2021MultilingualST,Han2021LearningSS,Dong2021ConsecutiveDF} has become popular, since it has lower latency and less error propagation compared with previous ST methods.
Recent studies have also shown that there is no significant difference between end-to-end models and cascaded systems in translation performance~\cite{Bentivogli2021CascadeVD}.

In many practical application scenarios, such as political negotiations, business meetings, etc., there is no available in-domain speech-translation dataset to conduct the end-to-end training, which essentially limits the promotion of E2E-ST systems.
The most common practice is that the E2E-ST model learns knowledge well enough in the general domain, and then it is directly used to translate speech input in the target domain.
Unfortunately, due to the domain shift issue~\cite{Gretton2006AKM,Ramponi2020NeuralUD}, the generalization capabilities of current end-to-end models are somehow weak across different scenarios.
Instead of speech-translation annotations, parallel text corpus in the target domain is usually abundant and easy to collect. 
Thus, it is essential to explore and extend the capability of E2E-ST systems in this scenario, in which a large amount of in-domain bilingual text is utilized.

In this paper, we focus on this domain adaptation setting and attempt to replace the domain-specific parameter updating in neural-based E2E-ST models with a non-parametric search to make it adaptable and achieve domain adaptation without any speech-translation annotations. 
Actually, the non-parametric approach $k$NN-MT, recently proposed by~\citet{Khandelwal2021NearestNM}, is a promising alternative to reach this goal.
The $k$NN-MT equips the pre-trained neural machine translation (NMT) model with a $k$-nearest-neighbor ($k$NN) classifier over an external datastore to improve translation accuracy without retraining.
However, it requires the in-domain speech-translation corpus to construct an effective datastore when we apply this method in the speech translation setting.
To tackle this problem, we propose a novel \textbf{N}on-\textbf{P}arametric \textbf{D}omain \textbf{A}daptation framework based on $k$NN-MT for E2E-ST, named as NPDA-$k$NN-ST.
Its key core is to directly leverage the in-domain text translation corpus to generate the corresponding datastore and encourage it to play a similar role as the real in-domain speech-translation data, through the carefully designed architecture and loss function.

Specifically, we first incorporate an additional trainable encoder for text modeling into the pre-trained E2E-ST model.
Based on that, we make the decoder's output representations for text and speech translation tasks close, through reducing the representation inconsistency of these two tasks in triplet training data and keeping the parameters of the original pre-trained E2E-ST model fixed.
In this way, the additional encoder module learns the semantic mapping in feature space from the source language text to the speech signal, which enables the construction of an effective in-domain datastore when text translation data is involved only.
Then we introduce a $k$NN classifier to produce the final translation distribution based on the domain-specific datastore built by the correspondent text translation data.
Meanwhile, the universal output representation is adopted to perform a similarity search and guides the translation process.

We evaluate our approach on the Europarl-ST benchmark and demonstrate that our method significantly outperforms the strong in-domain fine-tuning strategy by 3.85 BLEU scores on average in all translation directions when only using large-scale in-domain text translation data.
Additional experiments on Europarl-ST and MuST-C datasets verify that the in-domain text translation datastore generated by our method could play a similar role with the real in-domain speech-translation data, thanks to the universal output representation. 
\vspace{-5.5pt}

\section{Background}
\vspace{-2.5pt}
\subsection{End-to-End Speech Translation}
E2E-ST systems receive speech signals in a source language and directly generate the text in a target language without an intermediate transcription process.
Concretely, the E2E-ST corpus consists of a set of triplet data $\mathcal{D}_{ST}=\left\{(\mathbf{x}^{(n)}, \mathbf{z}^{(n)}, \mathbf{y}^{(n)})\right\}_{n=1}^{N}$, where $\mathbf{x}^{(n)}=(x_1^{(n)}, x_2^{(n)},...,x_{|\mathbf{x}^{(n)}|}^{(n)})$ is the input sequence of the speech wave (in most cases, acoustic features are used), $\mathbf{z}^{(n)}=(z_1^{(n)},z_2^{(n)},...,z_{|\mathbf{z}^{(n)}|}^{(n)})$ represents the transcription sequence from the source language and $\mathbf{y^{(n)}}=(y_1^{(n)}, y_2^{(n)},...,y_{|\mathbf{y}^{(n)}|}^{(n)})$ denotes the translation sequence of target language. 
The goal of E2E-ST model is to
seek an optimal translation sequence $\mathbf{y}$ without generating an intermediate transcription $\mathbf{z}$, and the standard training objective is to optimize the maximum likelihood estimation (MLE) of the training data $\mathcal{D}_{ST}$:
\begin{equation}
    \mathcal{L}_{ST}(\theta)=\frac{1}{N} \sum_{n=1}^{N} \log P\left(\mathbf{y^{(n)}}  \mid \mathbf{x^{(n)}} ; \theta\right), 
    \label{eq:vanilla_st}
\end{equation}
where a single encoder-decoder structure is adopted to fit the conditional distribution $P(\mathbf{y^{(n)}}|\mathbf{x^{(n)}})$ and $\theta$ is the model parameter. 
To develop high-quality E2E-ST systems, ASR and MT tasks ($(\mathbf{x}^{(n)}, \mathbf{z}^{(n)})$ and $(\mathbf{z}^{(n)}, \mathbf{y}^{(n)})$) are typically used to pre-train the encoder and decoder, respectively.~\cite{Bansal2019PretrainingOH, Wang2020CurriculumPF}.
However, in practice, it is not realistic to obtain a large amount of high-quality speech-translation data in every domain that we are interested in, while in-domain text translation corpus is usually cheaper and easier to collect.  
Thus, it is essential to investigate the capability of E2E-ST model that uses large-scale in-domain text translation corpus to achieve domain adaptation, making E2E-ST systems more practical.

\subsection{Nearest Neighbor Machine Translation}
Recently, $k$NN-MT~\cite{Khandelwal2021NearestNM} has shown the promising capability of directly augmenting the pre-trained NMT model with domain-specific token-level $k$NN retrieval to improve the translation performance without retraining.
$k$NN-MT mainly involves two steps: datastore creation and token inference with cached datastore.

\begin{figure*}[ht]
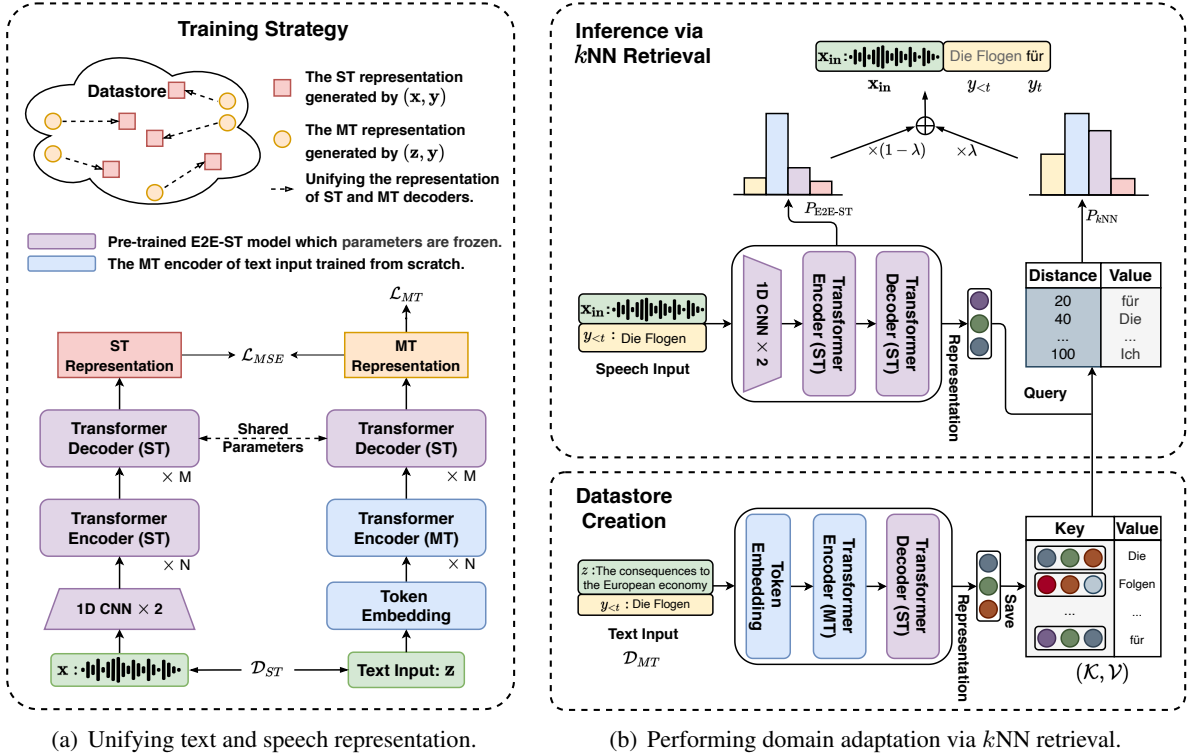

    \centering
    \subfigure[Unifying text and speech representation.]{
        \centering
        \includegraphics[height=9.4cm]{./method_training.pdf}
        \label{fig:npda-knn-st_model_training}
    }
    \subfigure[Performing domain adaptation via $k$NN retrieval.]{
        \centering
        \includegraphics[height=9.4cm]{./method_inference.pdf}
        \label{fig:npda-knn-st_model_inference}
    }
    \caption{The overview of our non-parametric domain adaptation framework for E2E-ST (NPDA-$k$NN-ST).}
    \label{fig:npda-knn-st_model}
\end{figure*}

\paragraph{Datastore Creation.}
The datastore of $k$NN-MT is the cache of a set of key-value pairs.
Given a parallel sentence pair $(\mathbf{z},\mathbf{y}) \in (\mathcal{Z}, \mathcal{Y})$, the pre-trained NMT model generates the context representation $f_\theta(\mathbf{z}, y_{\textless t})$ at each timestep $t$.
Then the whole datastore $(\mathcal{K},\mathcal{V})$ is constructed by taking the output hidden state $f_\theta(\mathbf{z}, y_{\textless t})$ as key and $y_t$ as value:
\begin{align}
(\mathcal{K}, \mathcal{V}) = \bigcup_{(\mathbf{z}, \mathbf{y}) \in (\mathcal{Z}, \mathcal{Y})} \{(f_\theta(\mathbf{z}, y_{\textless t}), y_t), \forall y_t \in \mathbf{y} \}. 
\end{align}

\paragraph{Inference via $k$NN Retrieval.} 
In the inference stage, $k$NN-MT model predicts the probability distribution of $t$-th target token $y_t$ with the context representation $f_\theta(\mathbf{z},y_{\textless t})$.
Specifically, $k$NN-MT utilizes the context representation to query the cached datastore $(\mathcal{K}, \mathcal{V})$ and retrieves $k$ nearest neighbor key-value pairs w.r.t. Euclidean distance. 
Then the probability distribution of $y_t$ generated by $k$NN retrieval is calculated as follow:
\begin{align}
    p_{k\text{NN}}&(y_t|\mathbf{z}, y_{\textless t}) \propto \label{eq:knn_prob} \\
&\sum_{(h_i, v_i)\in \mathcal{R}} \mathbb{1}_{y_t = v_i} \exp (\frac{-d(h_i, f_\theta(\mathbf{z}, y_{<t}))}{T}), \nonumber
\end{align}
where $\mathcal{R} = \{(h_i, v_i), i \in \{1, 2, ..., k\} \}$ is the set of $k$ nearest neighbors, $d(\cdot , \cdot)$ represents the squared Euclidean distance and $T$ is the temperature to control the sharpness of softmax function. 
The final output distribution is an interpolation between distributions from the NMT model and $k$NN retrieved neighbors with a tuned parameter $\lambda \in [0, 1]$:
\begin{equation}
\begin{split}
    p(y_t|\mathbf{z},y_{\textless t}) & = \lambda \ p_{\textrm{$k$NN}}(y_t|\mathbf{z},y_{\textless t}) \\
     & + (1-\lambda) \ p_{\textrm{NMT}}(y_t|\mathbf{z},y_{\textless t}). 
     \end{split}
\label{eq:final_prob}
\end{equation}

\section{Method}

When we apply $k$NN-MT in the speech translation task, it needs the real speech-translation corpus to build an effective datastore for $k$NN retrieval.
However, this requirement could not be satisfied in the domain adaptation scenario mentioned before, as there is no available in-domain speech-translation corpus.
In this paper, we focus on this setting and target to replace the domain-specific parameter updating with a non-parametric search to achieve domain adaptation. 
We design a novel \textbf{N}on-\textbf{P}arametric \textbf{D}omain \textbf{A}daptation framework based on $k$NN-MT for E2E-ST, named as NPDA-$k$NN-ST.
The overview framework of NPDA-$k$NN-ST is illustrated in Figure~\ref{fig:npda-knn-st_model}, which is mainly divided into two parts: a) unifying text and speech representation to enable datastore creation; b) performing domain adaptation through $k$NN retrieval. 
Next, we would introduce the model architecture, training objective and inference process in detail. 

\subsection{Unifying Text and Speech Representation}
The NPDA-$k$NN-ST aims to directly build an in-domain effective datastore with only text translation corpus, making it play a similar role with the real in-domain speech-translation data.
It means that whether word tokens or speech signals are treated as input, we should construct the universal output representation for them in a unified model.
As shown in Figure \ref{fig:npda-knn-st_model_training}, 
we introduce an additional transformer encoder and reuse the original transformer decoder of the pre-trained E2E-ST model for source text modeling.
In this way, we only increase a few parameters for our approach.

Based on this model structure, we further attempt to make the decoder's output representations for text and speech translation tasks close, by which the text translation data could be leveraged to build an effective in-domain datastore.  
We achieve this by leveraging out-of-domain triplet data $\mathcal{D}_{ST}$, which is also adopted to build the pre-trained E2E-ST model.
More specifically, given a triplet data point in the corpus $(\mathbf{x}, \mathbf{z}, \mathbf{y}) \in \mathcal{D}_{ST}$, the original E2E-ST model takes speech-translation pair $(\mathbf{x}, \mathbf{y})$ as input and generates output representation $f_{(\theta_e,\theta_d)}{(\mathbf{x};\ y_{<t})}$ for each target token $y_t$.
Meanwhile, with corresponding text translation pair $(\mathbf{z}, \mathbf{y})$, the model with an additional transformer encoder produces another representation for $y_t$, which can be denoted as $f_{(\theta_e',\theta_d)}{(\mathbf{z};\ y_{<t})}$.
Next, we take the end-to-end paradigm and directly update the introduced transformer encoder by minimizing the squared Euclidean distance of the two sets of decoder representations and optimizing the MLE loss of text translation pair:
\begin{align}
\mathcal{L}_{MSE}(\theta_e') & = \frac{1}{N}  \sum_{(\mathbf{x},\mathbf{z},\mathbf{y})\in \mathcal{D}_{ST}} (  \notag \\ 
   \frac{1}{|\mathbf{y}|}\sum_{t=1}^{|\mathbf{y}|} || & f_{(\theta_e, \theta_d)} {(\mathbf{x};y_{<t})} - f_{(\theta_e', \theta_d)}{(\mathbf{z}; y_{<t})} ||^2 ), \notag  \\
 \mathcal{L}_{MT}(\theta_e') & = \frac{1}{N} \sum_{n=1}^{N} \log P\left(\mathbf{y^{(n)}}  \mid \mathbf{z^{(n)}} ; \theta_e', \theta_d \right), \notag  \\
\mathcal{L}(\theta_e') & =\mathcal{L}_{MT}(\theta_e') - \mathcal{L}_{MSE}(\theta_e'),
\label{eq:mse}
\end{align}
where $\theta_e$ and $\theta_d$ are parameters of encoder and decoder in the pre-trained E2E-ST model respectively, $\theta_e'$ represents the parameter of the new transformer encoder and token embedding, and we keep $\theta_e$ and $\theta_d$ fixed during this training process to avoid the E2E-ST performance degradation in the inference stage.
The out-of-domain validation set and its correspondent loss are adopted to select the best model in our experiments. 

\subsection{Domain Adaptation via $k$NN Retrieval}

We consider the domain adaptation scenario of E2E-ST that only domain-specific text translation corpus $\mathcal{D}_{MT}=\left\{(\mathbf{z}^{(m)}, \mathbf{y}^{(m)})\right\}_{m=1}^{M}$ is available. 
During domain adaptation, the entire inference process is illustrated in Figure \ref{fig:npda-knn-st_model_inference}.
Once we gain the well-trained model with Equation \ref{eq:mse}, the new transformer encoder and original transformer decoder of pre-trained E2E-ST model are utilized to forward pass the text translation corpus $\mathcal{D}_{MT}$ to create an in-domain datastore $(\mathcal{K},\mathcal{V})$. 
This construction process is the same as the $k$NN-MT.
Due to the universal decoder's representation, this datastore is directly used for the $k$NN retrieval when translating in-domain speech input $\mathbf{x_{in}}$.
The final probability of NPDA-$k$NN-ST to predict the next token $y_{t}$ is an interpolation of two distributions with a tuned hyper-parameter $\lambda$:
\begin{equation}
\begin{split}
    p(y_t|\mathbf{x_{in}},y_{\textless t}) & = \lambda \ p_{\textrm{$k$NN}}(y_t|\mathbf{x_{in}},y_{\textless t}) \\
    + & ( 1  - \lambda) \ p_{\textrm{E2E-ST}}(y_t|\mathbf{x_{in}},y_{\textless t}),
\end{split}
\label{eq:knn-st-final_prob}
\end{equation}
where $p_{\textrm{E2E-ST}}$ indicates the general domain E2E-ST prediction and $p_{\textrm{kNN}}$ represents the in-domain retrieval based on Equation~\ref{eq:knn_prob}. 
Actually, this prediction way could also be substituted with other $k$NN variants~\cite{Zheng2021AdaptiveNN,He2021EfficientNN,Meng2021FastNN} to achieve better model performance or inference speed.

\section{Experiments}

\subsection{Setup}

We conduct experiments to evaluate our proposed approach in two aspects: a) domain adaptation on Europarl-ST benchmark with the E2E-ST model pre-trained on MuST-C dataset, and vice versa;
b) the performance comparisons on MuST-C benchmark when speech-translation and text-translation data are leveraged to build datastore, respectively.

\paragraph{MuST-C Dataset.} 
MuST-C~\cite{Gangi2019MuSTCAM} is a publicly large-scale multilingual ST corpus, which is built from English TED Talks and consists of triplet data: source speech, source transcription, and target translation.
It contains translations from English (EN) to 8 languages: Dutch (NL), French (FR), German (DE), Italian (IT), Portuguese (PT), Romanian (RO), Russian (RU) and Spanish (ES).
See Appendix~\ref{sec:appendix-dataset-statistics} for detailed statistics.

\paragraph{Europarl-ST Dataset.} 
Europarl-ST~\cite{IranzoSnchez2020EuroparlSTAM} collects from the official transcriptions and translations of European Parliament debate.
For domain adaptation, we select seven languages (DE, FR, IT, RO, NL, PT, and ES) that intersect with MuST-C.
The training size of Europarl-ST is one-ninth of MuST-C, and our method only leverages the bilingual text in the entire data to achieve domain adaptation.
The detailed statistics of the dataset are shown in Appendix~\ref{sec:appendix-dataset-statistics}.

\paragraph{Europarl-MT Dataset.} 
In order to further verify the performance of our proposed method with large-scale text translation data, we introduce the easily accessible in-domain parallel corpus -- Europarl-MT~\cite{Koehn2005EuroparlAP}. 
In our experiments, we randomly select 2M sentence pairs for each translation direction, except for the EN-RO translation direction. 
We adopt the entire Europarl-MT for EN-RO, which consists of almost 400k bilingual samples.

\paragraph{Baselines.}
We compare our proposed approach (NPDA-$k$NN-ST) with several baselines:
\begin{itemize}[leftmargin=*]
\setlength{\itemsep}{0pt}
\item \textbf{E2E-ST-Base}: The pre-trained E2E-ST model built by the MuST-C dataset. It is also treated as the pre-trained model for NPDA-$k$NN-ST.
\item \textbf{E2E-ST-SP}: The in-domain E2E-ST model on Europarl-ST. Its training process is consistent with E2E-ST-Base.
\item \textbf{E2E-ST-FT}: The fine-tuned version of E2E-ST-Base using the Europarl-ST corpus.
\item \textbf{LNA-D}: The multilingual E2E-ST model proposed by~\citet{Li2021MultilingualST}. It integrates Wave2vec 2.0~\cite{Baevski2020wav2vec2A} and mBART~\cite{Chipman2021mBARTMM}, while layernorm and attention layers in the decoder are fine-tuned with CoVoST dataset~\cite{Wang2020CoVoSTAD}. 
\item \textbf{$k$NN-MT}: We directly apply $k$NN-MT~\cite{Khandelwal2021NearestNM} for E2E-ST-Base and construct the cached datastore with the in-domain speech-translation data.
\item \textbf{Shallow Fusion}: We utilize Europarl-MT dataset to train in-domain language model (LM). During inference, we re-score hypotheses with the weighted sum of the scores by the E2E-ST-Base and LM models~\cite{gulcehre2015using}.
\item \textbf{E2E-JT-ST-MT}: The joint-training model with the MuST-C and Europarl-MT datasets, which adopts the same model structure as our method and all model parameters are tune-able. 
\end{itemize}

\paragraph{Dataset Pre-processing and Implementation Details.} 
We follow the \textsc{Fairseq} S2T~\cite{Wang2020FairseqSF} recipes to perform data pre-processing. 
For the speech data in Europarl-ST and MuST-C, we extract an 80-dimensional log-Mel filter bank as the acoustic feature. 
For the external text translation data, we delete the bilingual data in Europarl-MT that intersects with the validation/test sets of the Europarl-ST dataset.
Refer to Appendix~\ref{sec:appendix-data-processing} for dataset pre-processing details.
All experiments are implemented based on the \textsc{Fairseq}~\cite{Ott2019fairseqAF} toolkit. 
For the model structure of all baselines, it consists of two one-dimensional convolutional layers with a downsampling factor of 4, 12 transformer encoder layers and 6 transformer decoder layers. 
During training, we deploy the Adam optimizer~\cite{Kingma2015AdamAM} with a learning rate of 2e-3 and 10K warm-up updates to optimize model parameters. 
All models are trained with one Tesla-V100 GPU and we set patience to 5 to select the best checkpoint on the validation set. 
More implementation details can be found in Appendix~\ref{sec:appendix-imple-details} and~\ref{sec:appendix-datastore-statistics}.
In all experiments, we report the case-sensitive BLEU score~\cite{Papineni2002BleuAM} using sacreBLEU\footnote{https://github.com/mjpost/sacrebleu, with a configuration of 13a tokenizer, case-sensitiveness, and full punctuation}. Our code is open-sourced at \href{https://github.com/duyichao/NPDA-KNN-ST}{https://github.com/duyichao/NPDA-KNN-ST}.

\begin{table*}[t]
    \centering
    \small
    \normalsize
    \setlength{\tabcolsep}{1.0mm}{
    \scalebox{0.79}{
        \begin{tabular}{l|cccc|c|ccccccc|cc}
            \toprule
            \multirow{2}{*}{\textbf{Model}} & \multicolumn{4}{c|}{\textbf{Used Data}} & \multicolumn{1}{c|}{\textbf{Params. (M)}}  & \multicolumn{7}{c|}{\textbf{Target Language}} \\
            & \multicolumn{1}{c}{$\ \ $\textbf{MC}} & \textbf{EP-ST} & \textbf{EP-MT} & \textbf{Extra.} & \multicolumn{1}{c|}{\textbf{Tuned/Total}}  & \textbf{DE} & \textbf{FR} & \textbf{ES} & \textbf{NL} & \textbf{IT} & \textbf{RO} & \textbf{PT} & \textbf{Avg.} \\
            \midrule
            E2E-ST-Base  & \checkmark & $\times$ & $\times$ & $\times$ & 0.0/31.1 & 15.71 & 16.45 & 23.49 & 16.06 & 14.25 & 16.95 & 18.28 & 17.31  \\
            LNA-D & $\times$ & $\times$ & $\times$ & \checkmark & 384.8/793.0 & \underline{22.50} & 30.00 & 32.23 & /     & 21.50 & /     & \underline{28.40} & /     \\
            \midrule
            E2E-ST-SP & $\times$ & \checkmark & $\times$ & $\times$ & 31.1/31.1  & 16.20 & 24.52 & 26.00 & 19.50 & 18.35 & 20.62 & 21.34 & 20.93  \\
            E2E-ST-FT & \checkmark & \checkmark & $\times$ & $\times$  & 31.1/31.1   & 21.84 & 30.97 & \underline{32.25} & \underline{23.77} & \underline{23.36} & \underline{25.47} & 26.30 & \underline{26.28} \\
            $k$NN-MT  & \checkmark & \checkmark & $\times$ & $\times$ & 0.0/31.1  & 18.29 & 27.69 & 28.93 & 20.70 & 20.45 & 22.37 & 23.08 & 23.07   \\
            NPDA-$k$NN-ST & \checkmark & \checkmark & $\times$ & $\times$ & 17.1/48.1   & 18.76 & 27.73 & 29.01 & 20.79 & 20.54 & 23.54 & 23.54 & 23.42 \\
            \midrule
            
            Shallow Fusion & \checkmark & $\times$ & \checkmark & $\times$ & 6.8/37.9 & 17.72 & 22.66 & 26.25 & 20.12 & 18.77 & 20.58 & 21.59 & 21.67 \\
            E2E-JT-ST-MT & \checkmark & $\times$ & \checkmark & $\times$ & 48.1/48.1 & 22.10 & \underline{33.62} & 31.28 & 21.35 & 22.18 & 23.21 & 23.62 & 25.34  \\
            NPDA-$k$NN-ST$^+$ & \checkmark & $\times$ & \checkmark & $\times$ & 17.1/48.1  & \textbf{23.23}$^\uparrow$ & \textbf{35.26}$^\Uparrow$ & \textbf{33.71}$^\Uparrow$ & \textbf{27.71}$^\Uparrow$ & \textbf{33.76}$^\Uparrow$ & \textbf{28.29}$^\Uparrow$ & \textbf{28.96}$^\Uparrow$ & \textbf{30.13}  \\
            \bottomrule
            \end{tabular}
        }
    }  
    \caption{BLEU score [\%] of different methods on the Europarl-ST dataset. ``Tuned Params.'' refers to the number of fine-tuned parameters. 
    ``NPDA-$k$NN-ST$^+$'' directly uses large-scale Europarl-MT data to build the in-domain datastore, while ``NPDA-$k$NN-ST'' leverages the text translation part in the Europarl-ST training data. 
    ``MC, EP-ST, EP-MT and Extra.'' means whether the method uses MuST-C, Europarl-ST, Europarl-MT and external data, respectively. ``$^{\Uparrow/\uparrow}$'' indicates ``NPDA-$k$NN-ST$^{+}$'' significant difference ($p < 0.01/ 0.05$) from strong in-domain baseline ``E2E-ST-FT'', tested by bootstrap re-sampling~\cite{Koehn2004StatisticalST}.
    }
    \label{table:da_on_europarl}
\end{table*}

\begin{table*}[t]
    \small
    \centering
    \normalsize
    \setlength{\tabcolsep}{1.0mm}{
    \scalebox{0.85}{
        \begin{tabular}{l|ccc|c|ccccccc|ccc}
            \hline    
            \toprule
            \multirow{2}{*}{\textbf{Model}} & \multicolumn{3}{c|}{\textbf{Used Data}} & \multicolumn{1}{c|}{\textbf{Params. (M)}}  & \multicolumn{7}{c|}{\textbf{Target Language}} \\
            & \ \textbf{EP-ST} & \textbf{MC-ST} & \textbf{MC-MT} & \multicolumn{1}{c|}{\textbf{Tuned/Total}}  & \textbf{DE} & \textbf{FR} & \textbf{ES} & \textbf{NL} & \textbf{IT} & \textbf{RO} & \textbf{PT} & \textbf{Avg.} \\
            \midrule
            E2E-ST-SP      & \checkmark & $\times$ & $\times$   & 0.0/31.1  & 4.14 & 4.60 & 5.35 & 4.50 & 1.94 & 3.90 & 5.09 & 4.22   \\
            Shallow Fusion & \checkmark & $\times$ & \checkmark & 6.8/37.9  & 4.72 & 5.33 & 6.13 & 4.14 & 2.01 & 4.35 & 5.79 & 4.64 \\
            E2E-JT-ST-MT   & \checkmark & $\times$ & \checkmark & 48.1/48.1 & 5.73 & 7.58 & 7.62 & 5.85 & \textbf{3.83} & 5.05 & 6.45 & 6.02 \\
            \midrule
            $k$NN-MT       & \checkmark & \checkmark & $\times$ & 0.0/31.1   & \textbf{5.80}$^\Uparrow$ & 8.02$^\Uparrow$ & 8.21$^\Uparrow$ & 6.02$^\Uparrow$ & 3.41$^\Uparrow$ & \textbf{5.23}$^\Uparrow$ & 7.13$^\Uparrow$ & 6.26 \\
            NPDA-$k$NN-ST  & \checkmark& $\times$ & \checkmark  & 17.1/48.1 & 5.70$^\Uparrow$ & \textbf{8.24}$^\Uparrow$ & \textbf{8.28}$^\Uparrow$ & \textbf{6.19}$^\Uparrow$ & 3.45$^\Uparrow$ & \textbf{5.21}$^\Uparrow$ & \textbf{7.19}$^\Uparrow$ & \textbf{6.32} \\
            \bottomrule
            \hline
            \end{tabular}
        }
    }  
    \caption{BLEU score [\%] of different domain adaptation methods on the MuST-C dataset. ``MC-ST/MC-MT'' indicates whether the method uses MuST-C ST/MT data, respectively. ``$^{\Uparrow}$'' indicates the method significant difference ($p < 0.01$) from baseline ``E2E-ST-SP''.}
    \label{table:da_on_mustc}
\end{table*}

\subsection{Main Results}
\paragraph{Domain Adaptation on Europarl-ST.} 
We first evaluate the domain adaptation performance of NPDA-$k$NN-ST on Europarl-ST.
As illustrated in Table~\ref{table:da_on_europarl}, NPDA-$k$NN-ST obtains the significant improvements over E2E-ST-Base in all language pairs. 
Once the large-scale Europarl-MT data is involved, NPDA-$k$NN-ST$^+$ achieves 12.82 BLEU improvements over E2E-ST-Base on average, even significantly outperforming strong in-domain fine-tuning approach E2E-ST-FT.
Benefiting from the language model trained on large-scale in-domain data, Shallow Fusion gains similar performance to E2E-ST-SP, but it is still inferior to NPDA-$k$NN-ST.
E2E-JT-ST-MT achieves better performance by jointly training the entire model with large in-domain parallel text and out-of-domain speech data, but still falls short of NPDA-$k$NN-ST$^+$.
These results prove that our proposed method can make full use of in-domain parallel text to achieve domain adaptation when in-domain speech translation data is inaccessible.
In addition, NPDA-$k$NN-ST obtains comparable translation performance with $k$NN-MT that leverages the truly in-domain speech-translation data to construct a datastore. 
It further indicates that our method could generate an effective in-domain datastore with text translation data, which is equivalent to the real speech-translation data. 
We also compare our proposed method with LNA-D that builds the large multilingual E2E-ST model based on Wave2vec and mBART.
In spite of adopting a huge model scale and pre-training techniques, this approach still fails to outperform NPDA-$k$NN-ST$^+$ due to the domain shift problem.
This result shows the necessity of domain adaptation when applying large-scale general E2E-ST models in a certain domain.

\paragraph{Domain Adaptation on MuST-C.}
We further reverse the domain adaptation direction to verify the performance of our approach, such as domain adaptation to MuST-C using E2E-ST-SP. 
From Table~\ref{table:da_on_mustc}, we can see that NPDA-$k$NN-ST still significantly outperforms E2E-ST-SP and Shallow Fusion, yielding comparable results to E2E-JT-ST-MT.
Actually, Europarl-ST data is too small to build a good generic model, and its domain coverage is too narrow (i.e., only the political domain), resulting in the poor transfer performance of our method and low translation results of all methods.
It also brings an interesting research direction that incorporates our method with the large E2E-ST model, such as LNA-D, and we leave it as future work.

\begin{table*}[t]
    \small
    \normalsize
    \centering
    \setlength{\tabcolsep}{1.5mm}{
    \scalebox{0.88}{
        \begin{tabular}{c|l|c|cccccccc|c}
                \toprule
                & \multirow{2}{*}{\textbf{Model}}   & \multicolumn{1}{c|}{\textbf{Params. (M)}} & \multicolumn{8}{c|}{\textbf{Target Language}} \\
                & & \multicolumn{1}{c|}{\textbf{Tuned/Total}} & \textbf{DE} & \textbf{FR} & \textbf{ES} & \textbf{NL} & \textbf{IT} & \textbf{RO} & \textbf{PT} & \textbf{RU} & \textbf{Avg.} \\
                \midrule
                \multirow{4}{*}{\rotatebox[]{90}{\textbf{Bilingual}}} 
                & E2E-ST-Base &  31.1/31.1 & \underline{22.57} & \underline{32.61} & \underline{27.08} & \underline{27.46} & 22.74 & \underline{21.80} & \underline{28.07} & \underline{15.45} & \underline{24.72}    \\
                & AFS & - & 22.40 & 31.60 & 26.90 & 24.90 & \underline{23.00} & 21.00 & 26.30 & 14.70 & 23.85    \\
                \cmidrule {2 -12}
                & $k$NN-MT &   0.0/31.1 & 22.97$^\uparrow$  & 33.00$^\uparrow$  & 27.99$^\Uparrow$  & 27.93$^\uparrow$  & \textbf{23.55}$^\uparrow$  & 22.16  & 28.80$^\uparrow$  & 15.73$^\uparrow$  & 25.27 \\
                & NPDA-$k$NN-ST & 17.1/48.1 & \textbf{23.08}$^\uparrow$  & \textbf{33.24}$^\Uparrow$  & \textbf{28.03}$^\Uparrow$  & \textbf{28.11}$^\uparrow$  & 23.44$^\uparrow$  & \textbf{22.22}$^\uparrow$  & \textbf{28.83}$^\Uparrow$  & \textbf{15.82}$^\uparrow$  & \textbf{25.35}  \\
                \midrule
                \multirow{5}{*}{\rotatebox[]{90}{\textbf{Multilingual}}}
                & E2E-ST-Base &   76.3/76.3 & 24.18 & \underline{34.98} & 28.28 & \underline{28.80} & 24.62 & 23.22 & \underline{31.13} & 15.88 & 26.39  \\
                & LNA-D &  76.3/76.3 & 24.16 & 34.52 & 28.30 & 28.35 & 24.46 & 23.29 & 30.51 & 15.84 & 26.18   \\ 
                & Adapter Tuning  & 76.3/76.3 & \underline{24.63} & 34.75 & \underline{28.73} & \underline{28.80} & \underline{24.96} & \underline{23.70} & 30.96 & \underline{15.89} & \underline{26.61} \\ 
                \cmidrule {2 -12}
                & $k$NN-MT & 0.0/76.3 & 25.15$^\uparrow$  & \textbf{35.67}$^\Uparrow$  & \textbf{30.22}$^\Uparrow$  & \textbf{30.36}$^\Uparrow$  & 25.83$^\Uparrow$  & 23.66  & \textbf{31.67}$^\uparrow$  & 17.16$^\Uparrow$  & 27.47 \\
                & NPDA-$k$NN-ST &  23.7/100.0 & \textbf{25.21}$^\Uparrow$  & 35.56$^\uparrow$  & 30.05$^\Uparrow$  & 30.31$^\Uparrow$  & \textbf{25.91}$^\Uparrow$  & \textbf{23.90}$^\uparrow$  & 31.66$^\uparrow$  & \textbf{17.23}$^\Uparrow$  & \textbf{27.48} \\
                \bottomrule
        \end{tabular}
        }
    }  
    \caption{BLEU score [\%] of different E2E-ST methods on the MuST-C dataset. ``AFS'' and ``Adapter Tuning'' represent the methods proposed by \citet{Zhang2020AdaptiveFS} and \citet{Le2021LightweightAT}, respectively. Besides, \citet{Le2021LightweightAT} reproduce the translation performance of ``LNA-D'' on the MuST-C dataset for fair comparison. ``$^{\Uparrow/\uparrow}$'' indicates ``NPDA-$k$NN-ST/$k$NN-ST'' significant difference ($p < 0.01/ 0.05$) from the backbone ``E2E-ST-Base''.}
    \label{table:general_domain_results}
\end{table*}

\begin{table*}[t]
    \small
    \centering
    \setlength{\tabcolsep}{1.5mm}{
    \scalebox{1.02}{
        \begin{tabular}{l|l|ccccccc|cc}
                \toprule
                \textbf{Metric} & \textbf{Model} & \textbf{DE} & \textbf{FR} & \textbf{ES} & \textbf{NL} & \textbf{IT} & \textbf{RO} & \textbf{PT} & \textbf{Avg.} \\
                \midrule
                \multirow{3}{*}{\textbf{BLEU Score($\uparrow$)}} &
                NPDA-$k$NN-ST  & \textbf{18.76} & \textbf{27.73} & \textbf{29.01} & \textbf{20.79} & \textbf{20.54} & \textbf{23.54} & \textbf{23.54} & \textbf{23.42} \\
                & \ \ - w/o MSE Loss  & 18.44	& 26.66	& 28.10 & 19.93 & 19.89 & 22.20 & 22.45 & 22.52  \\
                & \ \ - Optimize Embedding Only  & 18.50 & 27.42 & 28.64 & 20.44 & 20.15 & 22.92 & 23.09 & 23.02 \\
                \midrule
                \multirow{3}{*}{\textbf{Cosine Similarity ($\uparrow$)}} &
                NPDA-$k$NN-ST  & \textbf{0.865} & \textbf{0.874}	& \textbf{0.858}	& \textbf{0.860} & \textbf{0.867} & \textbf{0.861} & \textbf{0.850} & \textbf{0.862} \\
                & \ \ - w/o MSE Loss  & 0.827 & 0.836 & 0.811 & 0.817 & 0.825 & 0.828 & 0.809 & 0.822   \\
                & \ \ - Optimize Embedding Only  & 0.844 & 0.857 & 0.839 & 0.844 & 0.849 & 0.845 & 0.832 & 0.844  \\
                \midrule
                \multirow{1.5}{*}{\textbf{Squared Euclidean}} &
                NPDA-$k$NN-ST & \textbf{5.387}	& \textbf{4.723} & \textbf{5.050}	& \textbf{5.637} & \textbf{5.098} & \textbf{4.996} & \textbf{5.707} & \textbf{5.228}  \\
                 \multirow{1.5}{*}{\textbf{Distance ($\downarrow$)}}
                & \ \ - w/o MSE Loss & 6.260 & 5.566 & 6.070 & 6.650 & 6.040 & 5.938 & 6.690 & 6.173  \\
                & \ \ - Optimize Embedding Only & 5.610 & 4.863 & 5.400 & 5.950 & 5.434 & 5.266 & 6.043 & 5.509   \\
                \bottomrule
            \end{tabular}
        }
    }  
    \caption{BLEU score [\%], cosine similarity and squared euclidean distance of our approach's variants on the Europarl-ST dataset. ``w/o MSE Loss'' means that the MSE loss function is removed. ``Optimize Embedding Only'' indicates that only the token embedding is introduced to the pre-trained E2E-ST model and fine-tuned.
    }
    \label{table:ablation-study}
\end{table*}

\paragraph{E2E-ST Performance on MuST-C.}
We investigate the effect of unifying text and speech representation with an additional encoder on MuST-C.
In this experiment, we compare the translation performance when speech and text translation data are leveraged to construct the datastore respectively, and verify the improvement of combining $k$NN retrieval with traditional E2E-ST models at the same time.
As illustrated in Table~\ref{table:general_domain_results}, we consider both bilingual and multilingual settings, and compare our method with other baselines, including AFS~\cite{Zhang2020AdaptiveFS}, LNA-D and Adapter Tuning~\cite{Le2021LightweightAT}.
When directly incorporating $k$NN retrieval into E2E-ST-Base, NPDA-$k$NN-ST yields 0.63 and 1.09 BLEU improvements on average in bilingual and multilingual settings, respectively.
These results indicate the benefit of introducing $k$NN retrieval, even when the E2E-ST model is trained on the same data.
In addition, NPDA-$k$NN-ST achieves similar performance with $k$NN-MT in both bilingual and multilingual settings, which proves the effectiveness of our proposed method on unifying text and speech representation again.

\subsection{Analysis}

\paragraph{Ablation Study.}
To analyze different modules in our method, we conduct an ablation study on the Europarl-ST dataset, including removing the MSE loss function and introducing only token embedding for unifying text and speech representation.
Except for the BLEU score, we measure the cosine similarity and squared euclidean distances between the synthetic representations generated by our method and ideals generated using ground-truth speech-translation data.
As shown in Table~\ref{table:ablation-study}, even without in-domain speech-translation data, NPDA-$k$NN-ST generates the representations that are close enough to the ideals (0.86 on cosine similarity and 5.2 on squared euclidean distances), leading to the efficient in-domain retrieval.
Two training losses, MSE and MLE, contribute significantly to the excellent performance of our approach.
Among that, the MT loss is more important, as optimizing the model with MSE loss only could not achieve effective domain adaptation.
Another observation is that our model could be smaller by introducing the token embedding and reusing the transformer encoder of the pre-trained E2E-ST model, causing small performance degradation.

\begin{figure*}[t]
    \centering
    \includegraphics[width = 16cm]{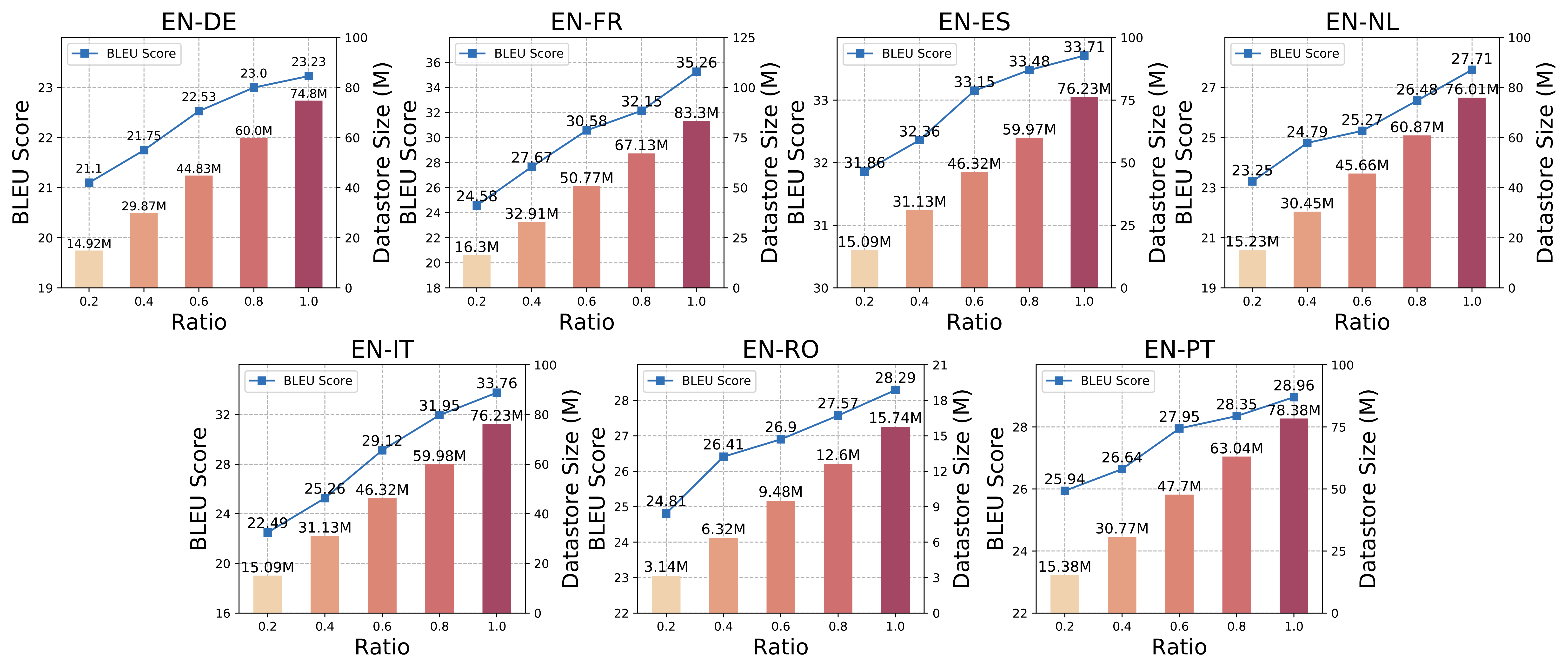}
    \caption{BLEU score [\%] of NPDA-$k$NN-ST with different datastore sizes on the Europarl-ST dataset.}
    \label{fig:ds-size}
\end{figure*}

\paragraph{The Impact of Datastore Size.}
As mentioned before, the datastore constructed by the bigger domain-specific text translation corpus seems to obtain better translation performance when using NPDA-$k$NN-ST. 
We investigate the performance differences caused by different datastore sizes on Europarl-ST.
For each translation direction, we adopt a ratio range of $(0.2, 0.4, 0.6, 0.8, 1.0)$ to randomly sample from Europarl-MT corpus to build the datastore of different scales for quick experiments. 
The detailed results are shown in Figure~\ref{fig:ds-size}.
In general, the translation performance in all directions is positively correlated with the datastore size.
More specifically, for EN-FR and EN-IT, model performance is increasing rapidly with the expansion of the datastore, exceeding 10 BLEU scores. 
The performance improvement in the DE, ES, NL and PT directions is relatively smooth. 
Since the overall datastore size of EN-RO is small, it still shows a reliable performance improvement. 
Thus, an enormous domain-specific text translation corpus further improves E2E-ST performance with NPDA-$k$NN-ST, but brings a larger datastore and slow inference speed, which is the trade-off in practice. 
Refer to Appendix~\ref{sec:appendix-inference-speed} for inference speed.

\section{Related Work}

\paragraph{Speech Translation.} 
Previous ST methods~\cite{Ney1999SpeechTC,Sperber2017NeuralLM, Zhang2019LatticeTF,Lam2021CascadedMW} cascade the ASR and MT tasks. 
With the rapid development of deep learning, the neural networks widely used  in ASR and MT have been adapted to construct a new end-to-end speech-to-text translation paradigm.
However, due to the scarcity of triplet training data, developing an E2E-ST model is still very challenging. 
Various techniques have been proposed to ease the training process by using source transcriptions,
including pre-training~\cite{Wang2020CurriculumPF}, multi-task learning~\cite{Weiss2017SequencetoSequenceMC,Anastasopoulos2018TiedML,Sperber2019AttentionPassingMF}, meta-learning~\cite{Indurthi2020EndendST}, interactive decoding~\cite{Liu2020SynchronousSR}, consecutive decoding~\cite{Dong2021ConsecutiveDF}, agreement-based training~\cite{Du2021RegularizingES} and adapter tuning~\cite{Le2021LightweightAT}. 
We first investigate the domain adaptation for E2E-ST and propose a non-parametric domain adaptation method to make the E2E-ST system more practical.

\paragraph{Domain Adaptation.} 
The domain adaptation approaches in MT field are mainly divided into two categories: 1) model-centric, which focuses on modifying the model architecture or the training objective to learn domain-related information~\cite{Wang2017InstanceWF,Wuebker2018CompactPM, Bapna2019SimpleSA, Guo2021ParameterEfficientTL};
2) data-centric, focusing on utilization of the monolingual corpus~\cite{Zhang2016ExploitingSM,Zhang2018JointTF}, synthetic corpus~\cite{Hu2019DomainAO,Wei2020IterativeDB}, or parallel corpus~\cite{Chu2017AnEC} in the specific domain for fine-tuning strategies. 
Recently, non-parametric methods provide a new paradigm for domain adaptation by retrieving the datastore of similar instances~\cite{Gu2018SearchEG,Khandelwal2021NearestNM,Zheng2021AdaptiveNN,zheng-etal-2021-non-parametric,He2021EfficientNN,Wang2021NonParametricOL}. 
We follow this research line and extend this non-parametric method in the domain adaptation scenario for E2E-ST.

\paragraph{Alignment of Speech and Text Representation.}
Recent research has shown that unified speech and text representations are helpful for downstream tasks~\cite{Chung2018UnsupervisedCA, Bapna2021SLAMAU,Akbari2021VATTTF,Tang2021ImprovingST}.
SLAM~\cite{Bapna2021SLAMAU} train a single encoder on large-scale text and speech data in a unsupervised manner, and further design corresponding speech-text alignment losses for downstream tasks.  
\citet{Tang2021ImprovingST} propose cross-attention regularization and online knowledge distillation to reduce the encoder representation differences between different modalities.
In this work, we make the decoder's output representation for ST and MT tasks close by reducing the inconsistency of their representation in the training triple data to enable the construction of a cross-modality datastore.

\section{Conclusion}
In this paper, we present a novel non-parametric method that leverages in-domain bilingual text to achieve domain adaptation for the E2E-ST system. 
This approach builds the universal output representation for text and speech translation tasks by a carefully designed architecture and loss function.
Based on that, a $k$NN classifier is introduced to improve translation performance with an external datastore constructed by the in-domain text translation data.
Experimental results demonstrate that our proposed method obtains significant improvement over pre-trained E2E-ST models when using large-scale in-domain bilingual text corpus. 
In the future, we would like to explore the combination of our method and the large-scale E2E-ST model, such as LNA-D.

\section*{Limitations}
The proposed approach constructs a datastore using text translation data from the target domain and utilizes $k$NN retrieval to assist pre-trained E2E-ST models for domain adaptation.
Our approach achieves a significant performance improvement over the basic model, but also brings time and space costs, i.e., storage overhead for datastore and time costs for $k$NN retrieval.
In practice, these costs are acceptable since we adopt \textsc{Faiss} to speed up $k$NN retrieval and reduce the storage requirement (as shown in Table~\ref{table:inference-speed2}).
We also encourage future work to further investigate how to build a smaller datastore as well as improve the efficiency of $k$NN retrieval.
Since the promising domain adaptation performance of our approach benefits from the strong foundation model, another interesting direction is to explore the combination of our method and the large-scale E2E-ST model, such as LNA-D.

\section*{Acknowledgements}
This work was supported by the grants from National Natural Science Foundation of China (No.U20A20229, 62072423), CAAI-Huawei MindSpore Open Fund (CAAIXSJLJJ-2021-007B), and the USTC Research Funds of the Double First-Class Initiative (No.YD2150002009). We appreciate Linan Yue, Yanqing An and Li Wang for the fruitful discussions. We thank the anonymous reviewers for helpful feedback on early versions of this work.
\bibliography{custom}
\bibliographystyle{acl_natbib}

\clearpage
\appendix
\section{Appendix}
\label{sec:appendix}

\subsection{Dataset Statistics}
\label{sec:appendix-dataset-statistics}
The statistics of MuST-C and Europarl-ST datasets are shown in Table~\ref{tab:must-c_stat} and~\ref{tab:Europarl-ST_stat}.

\subsection{Dataset Preprocessing} 
\label{sec:appendix-data-processing}
We follow the \textsc{Fairseq} S2T~\cite{Wang2020FairseqSF} recipes to perform data pre-processing. 
For speech data, both in Europarl-ST and MuST-C, acoustic features are 80-dimensional log-mel filter banks extracted with a stepsize of 10ms and a window size of 25ms. 
The acoustic features are normalized by global channel mean and variance.
The SpecAugment method~\cite{Park2019SpecAugmentAS} is used in all experiments and we remove samples consisting of more than 3k frames.
For external text translation data, we delete the bilingual data in Europarl-MT that intersects with validation/test sets of the Europarl-ST dataset.
We adopt unigram sentencepiece\footnote{https://github.com/google/sentencepiece} to build 5K and 8K sub-word vocabularies for the transcriptions and the translations, respectively.
For the multilingual model, both vocabulary sizes are set to 10K.
In all experiments, the MuST-C dataset is only used to construct the sub-word dictionary.

\subsection{Implementation Details}
\label{sec:appendix-imple-details}
All experiments are implemented based on the \textsc{Fairseq}\footnote{https://github.com/pytorch/fairseq}~\cite{Ott2019fairseqAF} toolkit. 
For the model structure of all baselines, it consists of two one-dimensional convolutional layers with a downsampling factor of 4, 12 transformer encoder layers, and 6 transformer decoder layers. 
The additional encoder in our approach includes 12 transformer encoder layers and token embedding, and all parameters are initialized randomly.
The input embedding size of the transformer layer is 256, the FFN layer dimension is 1024, and the number of self-attention heads is 4. 
We adopt 6 transformer decoder layers and the same parameters for LM training.
For the multilingual model, the above parameters are set to 512, 2048 and 8 respectively. 
During training, we deploy the Adam optimizer~\cite{Kingma2015AdamAM} with a learning rate of 2e-3 and 10K warm-up updates to optimize model parameters. 
Both label smoothing coefficient and dropout rate are set to 0.1. 
The batch size is set to 20K tokens, and we accumulate the gradient for every 4 batches. 
We train all models with one Tesla-V100 GPU and set patience to 5 to select the best checkpoint on the validation set. 
The \textsc{Faiss}\footnote{https://github.com/facebookresearch/faiss}~\cite{Johnson2021BillionScaleSS} is leveraged to construct the in-domain datastore and carry out fast nearest neighbor search.
We utilize the \textsc{Faiss} to learn $8192$ cluster centroids for each translation direction.
During inference, the beam size and length penalty are set to 5 and 0.6 for all methods and we search $64$ clusters for each target token when using \textsc{Faiss}. 
The performance of $k$NN-MT and NPDA-$k$NN-ST is highly related to the choice of hyper-parameters.
The hyper-parameters ($k$, $\lambda$ and $T$) for $k$NN retrieval are tuned on the in-domain validation set. 
We adopt grid search of $k\in\{4, 8, 16, 32\}$, $\lambda \in \{0.1, 0.2, ..., 0.9\}$ and $T\in\{1, 10, 20, 50, 100, 200 \}$ for each translation direction on Europarl-ST/MuST-C validation sets when using $k$NN-MT and NPDA-$k$NN-ST. 
The optimal choices of different datasets are shown in Table~\ref{table:hp-each-td-ep-st},~\ref{table:hp-each-td-ep2mustc} and~\ref{table:hp-each-td-mustc}.

\begin{table}[t]
    \centering
    \normalsize
    \scalebox{0.90}{
        \begin{tabular}{c|cccccc} 
            \toprule
            & \textbf{Speech} & \textbf{Train} & \textbf{Dev}  & \textbf{Test}  \\
                                           & \textbf{Duration}  & \textbf{Pairs} & \textbf{Pairs} & \textbf{Pairs} \\
            \midrule
            \textbf{DE}     & 408$\mathrm{~hrs}$ & 225,278 & 1,419 & 2,588 \\
            \textbf{FR}     & 492$\mathrm{~hrs}$ & 269,256 & 1,409 & 2,579 \\
            \textbf{ES}     & 504$\mathrm{~hrs}$ & 260,050 & 1,313 & 2,450 \\
            \textbf{IT}     & 465$\mathrm{~hrs}$ & 248,155 & 1,305 & 2,521 \\
            \textbf{NL}     & 442$\mathrm{~hrs}$ & 243,516 & 1,419 & 2,563 \\
            \textbf{PT}     & 385$\mathrm{~hrs}$ & 201,462 & 1,365 & 2,449 \\
            \textbf{RO}     & 432$\mathrm{~hrs}$ & 231,471 & 1,366 & 2,503 \\
            \textbf{RU}     & 489$\mathrm{~hrs}$ & 259,531 & 1,313 & 2,460 \\
            \bottomrule
        \end{tabular}
    }
    \caption{The statistics of all EN-X translation directions in the MuST-C dataset.}
    \label{tab:must-c_stat}
\end{table}

\begin{table}[t]
    \centering
    \normalsize
    \scalebox{0.90}{
        \begin{tabular}{c|cccccc} 
            \toprule
            & \textbf{Speech} & \textbf{Train} & \textbf{Dev}  & \textbf{Test}  \\
            & \textbf{Duration}  & \textbf{Pairs} & \textbf{Pairs} & \textbf{Pairs} \\
            \midrule
            \textbf{DE}     & 83$\mathrm{~hrs}$ & 32,629 & 1,321 & 1,254 \\
            \textbf{FR}     & 81$\mathrm{~hrs}$ & 31,778 & 1,282 & 1,215 \\
            \textbf{ES}     & 81$\mathrm{~hrs}$ & 31,608 & 1,273 & 1,268 \\
            \textbf{IT}     & 80$\mathrm{~hrs}$ & 29,553 & 1,123 & 1,131 \\
            \textbf{NL}     & 80$\mathrm{~hrs}$ & 31,402 & 1,270 & 1,236 \\
            \textbf{PT}     & 81$\mathrm{~hrs}$ & 31,751 & 1,295 & 1,263 \\
            \textbf{RO}     & 72$\mathrm{~hrs}$ & 28,599 & 1,071 & 1,096 \\
            \bottomrule
        \end{tabular}
    }
    \caption{The statistics of all EN-X translation directions in the Europarl-ST dataset.}
    \label{tab:Europarl-ST_stat}
\end{table}

\begin{table}[htb]
    \small
    \centering
    \normalsize
    \scalebox{0.85}{
        \begin{tabular}{c|cccccccccc}
                \toprule
                & \textbf{DE} & \textbf{FR} & \textbf{ES} & \textbf{NL} & \textbf{IT} & \textbf{RO} & \textbf{PT} \\
                \midrule
                \multicolumn{8}{c}{\textbf{$k$NN-MT}} \\
                \midrule
                \bm{$k$}        & 16  & 16  & 16  & 16  & 16  & 8   & 8   \\
                \bm{$\lambda$}  & 0.5 & 0.7 & 0.6 & 0.6 & 0.7 & 0.6 & 0.6 \\
                \bm{$T$}        & 10  & 20  & 10  & 20  & 20  & 50  & 50  \\
                \midrule
                \multicolumn{8}{c}{\textbf{NPDA-$k$NN-ST}} \\
                \midrule
                \bm{$k$}        & 16  & 32  & 16  & 16  & 32  & 16  & 32  \\
                \bm{$\lambda$}  & 0.5 & 0.7 & 0.6 & 0.7 & 0.7 & 0.7 & 0.7 \\
                \bm{$T$}        & 10  & 10  & 10  & 20  & 10  & 10  & 10  \\
                \midrule
                \multicolumn{8}{c}{\textbf{NPDA-$k$NN-ST$^+$}} \\
                \midrule
                \bm{$k$}        & 32  & 4   & 8   & 8   & 4   & 8   & 8   \\
                \bm{$\lambda$}  & 0.8 & 0.8 & 0.8 & 0.8 & 0.8 & 0.8 & 0.8 \\
                \bm{$T$}        & 10  & 10  & 10  & 10  & 10  & 10  & 10  \\
                \bottomrule
            \end{tabular}
    }
    \caption{The optimal choice of hyper-parameters for all EN-X translation directions on Europarl-ST validation set in domain adaptation experiments.}
    \label{table:hp-each-td-ep-st}
\end{table}

\begin{table}[htb]
    \small
    \centering
    \normalsize
    \setlength{\tabcolsep}{1.8mm}{
    \scalebox{0.85}{
        \begin{tabular}{c|cccccccccc}
                \toprule
                & \textbf{DE} & \textbf{FR} & \textbf{ES} & \textbf{NL} & \textbf{IT} & \textbf{RO} & \textbf{PT} \\
                \midrule
                \multicolumn{8}{c}{\textbf{$k$NN-MT}} \\
                \midrule
                \bm{$k$}       & 8   & 16  & 8   & 16  & 16  & 16  & 16         \\
                \bm{$\lambda$} & 0.5 & 0.5 & 0.6 & 0.6 & 0.8 & 0.1 & 0.7     \\
                \bm{$T$}       & 200 & 20  & 50  & 50  & 100 & 50  & 50            \\
                \midrule
                \multicolumn{8}{c}{\textbf{NPDA-$k$NN-ST}} \\
                \midrule
                \bm{$k$}       & 16  & 32  & 16  & 16  & 8   & 32  & 32              \\
                \bm{$\lambda$} & 0.6 & 0.6 & 0.6 & 0.7 & 0.5 & 0.3 & 0.2        \\
                \bm{$T$}       & 100 & 20  & 20  & 100 & 100 & 10  & 20              \\
                \bottomrule
            \end{tabular}
        }
    }  
    \caption{The optimal choice of hyper-parameters for all EN-X translation directions on MuST-C validation set in domain adaptation experiments.}
    \label{table:hp-each-td-ep2mustc}
\end{table}

\begin{table}[htb]
    \small
    \centering
    \normalsize
    \setlength{\tabcolsep}{1.8mm}{
    \scalebox{0.85}{
        \begin{tabular}{c|ccccccccccc}
                \toprule
                & \textbf{DE} & \textbf{FR} & \textbf{ES} & \textbf{NL} & \textbf{IT} & \textbf{RO} & \textbf{PT} & \textbf{RU} \\
                \midrule
                \multicolumn{8}{c}{\textbf{$k$NN-MT}} \\
                \midrule
                \bm{$k$} & 8 & 16 & 8 & 16 & 16 & 16 & 16 & 8         \\
                \bm{$\lambda$} & 0.2 & 0.3 & 0.2 & 0.2 & 0.3 & 0.1 & 0.2 & 0.3     \\
                \bm{$T$} & 10 & 20 & 20 & 50 & 10 & 50 & 10 & 20           \\
                \midrule
                \multicolumn{8}{c}{\textbf{NPDA-$k$NN-ST}} \\
                \midrule
                \bm{$k$} & 32 & 16 & 8 & 16 & 8 & 32 & 16 & 8             \\
                \bm{$\lambda$} & 0.4 & 0.3 & 0.3 & 0.3 & 0.3 & 0.3 & 0.2 & 0.3        \\
                \bm{$T$} & 10 & 20 & 20 & 50 & 10 & 10 & 20 & 20             \\
                \bottomrule
            \end{tabular}
        }
    }  
    \caption{The optimal choice of hyper-parameters for all EN-X translation directions on MuST-C validation set in E2E-ST experiments.}
    \label{table:hp-each-td-mustc}
\end{table}

\begin{table*}[t] 
    \small
    \centering
    \normalsize
    \scalebox{0.73}{
    \begin{tabular}{l|c|cccccccccccc}
        \toprule
        & & \textbf{DE} & \textbf{FR} & \textbf{ES} & \textbf{NL} & \textbf{IT} & \textbf{RO} & \textbf{PT}    \\ 
        \midrule
        \multirow{3}{*}{NPDA-$k$NN-ST} & $(\mathcal{K},\mathcal{V})$ 
                                & 1,220,631 & 1,265,862   & 1,160,737 & 1,153,677 & 1,139,009 & 1,083,567 & 1,194,161 \\ 
        & Datastore             & 597 MB    & 619 MB      & 568 MB    & 568 MB    & 557 MB    & 530 MB    & 584 MB    \\ 
        & Faiss index           & 93 MB     & 96 MB       & 89 MB     & 89 MB     & 87 MB     & 83 MB     & 91 MB     \\
        \midrule
        \multirow{3}{*}{NPDA-$k$NN-ST$^+$} & $(\mathcal{K},\mathcal{V})$ 
                            & 74,795,371 & 83,303,733  & 76,226,723 & 76,011,171 & 75,981,836 & 15,738,321 & 78,375,866 \\ 
        & Datastore         & 36 GB      & 40 GB       & 37 GB      & 37 GB      & 37 GB      & 7.6 GB     & 38 GB      \\ 
        & Faiss index       & 3.1 GB     & 3.5 GB      & 3.2 GB     & 3.1 GB     & 3.1 GB     & 224 MB     & 3.2 GB     \\
        \bottomrule
    \end{tabular}
    }
    \caption{The statistics of datastore for all EN-X translation directions on Europarl-ST dataset.}
    \label{table:disk-space-europarl}
\end{table*}

\begin{table*}[t] 
    \small
    \centering
    \normalsize
    \scalebox{0.73}{
    \begin{tabular}{l|c|cccccccccccc}
        \toprule
        & & \textbf{DE} & \textbf{FR} & \textbf{ES} & \textbf{NL} & \textbf{IT} & \textbf{RO} & \textbf{PT}  & \textbf{RU}   \\ 
        \midrule
        \multirow{3}{*}{\shortstack{NPDA-$k$NN-ST\\(Bilingual)}} & $(\mathcal{K},\mathcal{V})$ 
                                & 5,909,910 & 7,843,906   & 7,028,102 & 6,006,360 & 6,591,640 & 6,339,525 & 5,345,744 & 7,150,960 \\ 
        & Datastore             & 2.9 GB    & 3.8 GB      & 3.4 GB    & 2.9 GB    & 3.2 GB    & 3.1 GB    & 2.6 GB    & 3.5 GB \\ 
        & Faiss index           & 415 MB    & 547 MB      & 491 MB    & 421 MB    & 461 MB    & 444 MB    & 376 MB    & 500 MB \\
        \midrule
        \multirow{3}{*}{\shortstack{NPDA-$k$NN-ST\\(Multilingual)}} & $(\mathcal{K},\mathcal{V})$ 
                            & 7,587,793 & 9,530,628   & 8,507,191 & 7,572,305 & 8,121,129 & 7,819,137 & 6,626,153 & 9,460,741\\ 
        & Datastore         & 7.3 GB    & 9.1 GB      & 8.2 GB    & 7.3 GB    & 7.8 GB    & 7.5 GB     & 6.4 GB   & 9.1 GB \\ 
        & Faiss index       & 538 MB    & 671 MB      & 601 MB    & 537 MB    & 575 MB    & 554 MB     & 472 MB   & 666 MB\\
        \bottomrule
    \end{tabular}
    }
    \caption{The statistics of datastore for all EN-X translation directions on MuST-C dataset. }
    \label{table:disk-space-mustc}
\end{table*}

\subsection{Statistics of Datastore}
\label{sec:appendix-datastore-statistics}
The statistics of datastore used in our experiments are shown in Table~\ref{table:disk-space-europarl} and~\ref{table:disk-space-mustc}. 
Note that the datastore statistics of $k$NN-MT are exactly the same as those of NPDA-$k$NN-ST, due to the same number of ground truth tokens when building datastores.

\begin{table*}[!htb]
    \small
    \normalsize
    \centering
    \setlength{\tabcolsep}{1.0mm}{
    \scalebox{0.84}{
        \begin{tabular}{l|cccc|c|ccccccc|cc}
            \toprule
            \multirow{2}{*}{\textbf{Model}} & \multicolumn{4}{c|}{\textbf{Used Data}} & \multicolumn{1}{c|}{\textbf{Params. (M)}}  & \multicolumn{7}{c|}{\textbf{Target Language}} \\
            & \multicolumn{1}{c}{$\ \ $\textbf{MC}} & \textbf{EP-ST} & \textbf{EP-MT} & \textbf{Extra.} & \multicolumn{1}{c|}{\textbf{Tuned/Total}}  & \textbf{DE} & \textbf{FR} & \textbf{ES} & \textbf{NL} & \textbf{IT} & \textbf{RO} & \textbf{PT} & \textbf{Avg.} \\
            \midrule
            E2E-ST-Base  & \checkmark & $\times$ & $\times$ & $\times$ & 0.0/31.1 & 15.71 & 16.45 & 23.49 & 16.06 & 14.25 & 16.95 & 18.28 & 17.31  \\
            Cascade-SP & $\times$& \checkmark & $\times$ & \checkmark & / & 22.40 & 23.40 & 28.00 & / & / & / & / & /  \\
            \midrule
            Shallow Fusion & \checkmark & $\times$ & \checkmark & $\times$ & 6.8/37.9 & 17.72 & 22.66 & 26.25 & 20.12 & 18.77 & 20.58 & 21.59 & 21.67 \\
            E2E-JT-ST-MT & \checkmark & $\times$ & \checkmark & $\times$ & 48.1/48.1 & 22.10 & 33.62 & 31.28 & 21.35 & 22.18 & 23.21 & 23.62 & 25.34  \\
            Cascade-ST & \checkmark & $\times$ & \checkmark & $\times$ & 67.7/67.7 & 22.63 & 34.58 & 33.08 & 25.43 & 26.05 & 26.60 & 26.29 & 27.81  \\
            Cascade-ST$^*$ & \checkmark & $\times$ & \checkmark & $\times$ & 88.4/88.4 & \underline{24.71} & 34.60 & 33.70 & 26.55 & 29.94 & 26.38 & 30.18 & 29.44 \\
            NPDA-$k$NN-ST$^+$ & \checkmark & $\times$ & \checkmark & $\times$ & 17.1/48.1  & 23.23 & \underline{35.26} & \underline{33.71} & \underline{27.71} & \underline{33.76} & \underline{28.29} & \underline{28.96} & \underline{30.13}  \\
            NPDA-$k$NN-ST$^{++}$  & \checkmark & \checkmark & \checkmark & $\times$ & 48.1/48.1   & \textbf{24.90} & \textbf{35.95} & \textbf{34.32} & \textbf{28.44} & \textbf{34.69} & \textbf{29.68} & \textbf{32.83} & \textbf{31.54} \\

            \bottomrule
            \end{tabular}
        }
    }  
    \caption{BLEU score [\%] of different methods on the Europarl-ST dataset. ``Cascade-SP '' is the cascade model built by~\citet{IranzoSnchez2020EuroparlSTAM}. We also reproduce the performance of cascade methods with two dictionary settings, in which ``Cascade-ST'' adopts the same dictionary as our method and `Cascade-ST$^*$'' constructs the dictionary with both MuST-C and Europarl-MT datasets. }
    \label{table:da_cascade}
\end{table*}

\begin{table*}[!htb] 
    \centering
    \small
    \normalsize
    \scalebox{0.86}{
        \begin{tabular}{l|cc|c|ccccccc}
            \toprule
            \multirow{2}{*}{\textbf{Model}} & \multicolumn{2}{c|}{\textbf{Hard Disk Space}}  & \multirow{2}{*}{\bm{$k$}} & \multicolumn{4}{c}{\textbf{Inference Speed} (ms/sentence)} \\    
            & Datastore & Faiss Index & & batch=1 & batch=8 & batch=16 & batch=32   \\ 
            \midrule
            \multirow{1}{*}{E2E-ST-Base} & - & -  &  $0$    & 347.8 & 64.7 & 37.1 & 21.9  \\
            \midrule
            \multirow{1}{*}{E2E-ST-FT} 
             & - & -  &  $0$ & 349.5 $(\times 1.00)$ & 63.7 $(\times 0.98)$ & 37.5 $(\times 1.01)$ & 21.5 $(\times 0.98)$  \\
            \midrule
            \multirow{1}{*}{Cascade-ST}  & - & -  &  $0$ & 690.3 $(\times 1.98)$ & 127.3 $(\times 1.97)$ & 68.8 $(\times 1.85)$ & 40.1 $(\times 1.83)$  \\ 
            \multirow{1}{*}{Cascade-ST*} & - & -  &  $0$ & 788.0 $(\times 2.27)$ & 142.6 $(\times 2.20)$ & 80.1 $(\times 2.16)$ & 46.2 $(\times 2.11)$ \\ 
            \midrule
            \multirow{4}{*}{NPDA-$k$NN-ST} & \multirow{4}{*}{597 MB}  & \multirow{4}{*}{93 MB}
             & $4$        & 375.8 $(\times 1.08)$ & 70.5 $(\times 1.09)$ & 40.3 $(\times 1.09)$ & 24.5 $(\times 1.12)$  \\
             & & & $8$    & 379.2 $(\times 1.09)$ & 70.9 $(\times 1.10)$ & 41.4 $(\times 1.12)$ & 24.7 $(\times 1.13)$  \\
             & & & $16$   & 381.4 $(\times 1.10)$ & 70.1 $(\times 1.08)$ & 40.8 $(\times 1.10)$ & 24.2 $(\times 1.11)$  \\
             & & & $32$   & 374.9 $(\times 1.08)$ & 71.5 $(\times 1.11)$ & 39.8 $(\times 1.07)$ & 24.2 $(\times 1.11)$  \\
            \midrule
            \multirow{4}{*}{NPDA-$k$NN-ST$^+$}  & \multirow{4}{*}{36 GB}  & \multirow{4}{*}{3.1 GB}
             & $4$       & 419.3 $(\times 1.21)$ & 94.1 $(\times 1.45)$ & 63.4 $(\times 1.71)$ & 46.9 $(\times 2.14)$  \\
             & & & $8$   & 419.6 $(\times 1.21)$ & 96.3 $(\times 1.49)$ & 64.7 $(\times 1.74)$ & 46.9 $(\times 2.14)$  \\
             & & & $16$  & 420.6 $(\times 1.21)$ & 94.3 $(\times 1.46)$ & 63.6 $(\times 1.71)$ & 47.0 $(\times 2.15)$  \\
             & & & $32$  & 420.2 $(\times 1.21)$ & 93.3 $(\times 1.44)$ & 64.0 $(\times 1.73)$ & 46.9 $(\times 2.14)$  \\
            \bottomrule
        \end{tabular}
    }
    \caption{Inference speed of different methods in EN-DE direction of Europael-ST. 
    All results are the average of three runs on a server with 96-core  Intel(R) Xeon(R) Platinum 8255C CPU @ 2.50GHz and Tesla V100-SXM2-32GB GPU.}
    \label{table:inference-speed2}
\end{table*}

\subsection{Comparison with Cascade Methods}
Table~\ref{table:da_cascade} shows the performance comparisons of NPDA-$k$NN-ST$^+$ with different cascade systems, including Cascade-SP, Cascade-ST and Cascade-ST$^*$.
The Cascade-SP is built by~\citet{IranzoSnchez2020EuroparlSTAM} and we further reproduce cascade methods with two dictionary settings.
Cascade-ST adopts the same dictionary as NPDA-$k$NN-ST$^+$, while Cascade-ST$^*$ constructs a 40K byte-pair dictionary with MuST-C and Europarl-MT datasets for NMT model.
Both Cascade-ST and Cascade-ST$^*$ adopt the same ASR model that is trained on the MuST-C dataset.
The model structure of NMT in these two cascade systems contains a 6-layer transformer encoder and a 6-layer transformer decoder, in which input dimension, FFN layer dimension and attention heads are 512, 1024 and 4 respectively. 
We can observe that NPDA-$k$NN-ST$^+$ outperforms Cascade-ST in all translation directions, and the inference speed is more competitive (see Table~\ref{table:inference-speed2}). 
Cascade-ST$^*$ obtains significant improvement over Cascade-ST thanks to the better dictionary built on both MuST-C and Europarl-MT datasets. 
We believe that our proposed method could also benefit from such dictionary, but it requires leveraging such a dictionary to train the E2E-ST model at the beginning.

Intuitively, introducing in-domain triplet data could yield a better contextual representation for our method, which may help $k$NN to retrieve more accurate candidates and improve the final performance. We directly apply our method for the E2E-ST-FT model, in which we use in-domain data (Europarl-ST) to build the aligning representation (named NPDA-kNN-ST++).We conduct the experiments in the En-X translation directions, and the results are illustrated in Table~\ref{table:da_cascade}. As we expected, NPDA-kNN-ST++ achieves better performance than NPDA-kNN-ST+.

\subsection{Inference Speed Comparison}
\label{sec:appendix-inference-speed}
We compare the inference speed of four methods (E2E-ST-Base, E2E-ST-FT, NPDA-$k$NN-ST and NPDA-$k$NN-ST+) on Europael-ST EN-DE test set with different hyper-parameters ($k = 4, 8, 16, 32$) and batch sizes ($\text{batch} = 1, 8, 16, 32$).
As shown in Table~\ref{table:inference-speed2}, the inference time of NPDA-$k$NN-ST increases with the bigger datastore size.
The larger datastore means that more keys need to be retrieved during the inference phase, which reduces the inference speed. 
Nonetheless, when in-domain speech translation data is inaccessible to fine-tune the E2E-ST-Base model, it is still worth sacrificing part of the time and storage for higher performance.
Note that NPDA-$k$NN-ST only needs to load the Faiss index to perform $k$NN  retrieval and we could further replace the prediction way used in our paper with other $k$NN variants~\cite{He2021EfficientNN,Meng2021FastNN} to reduce the inference time.

\end{document}